# A Temporal Bayesian Network for Diagnosis and Prediction

**Gustavo Arroyo-Figueroa**
IIE – USP
AP 475-1, Cuernavaca, Morelos
62001 MÉXICO
garroyo@iie.org.mx

**Luis Enrique Sucar**
ITESM – Campus Morelos
AP C-99, Cuernavaca, Morelos
62020 MÉXICO
esucar@campus.mor.itesm.mx

## Abstract

Diagnosis and prediction in some domains, like medical and industrial diagnosis, require a representation that combines uncertainty management and temporal reasoning. Based on the fact that in many cases there are few state changes in the temporal range of interest, we propose a novel representation called Temporal Nodes Bayesian Network (TNBN). In a TNBN each node represents an event or state change of a variable, and an arc corresponds to a causal-temporal relation. The temporal intervals can differ in number and size for each temporal node, so this allows multiple granularity. Our approach is contrasted with a dynamic Bayesian network for a simple medical example. An empirical evaluation is presented for a more complex problem, a subsystem of a fossil power plant, in which this approach is used for fault diagnosis and event prediction with good results.

## 1. INTRODUCTION

Artificial intelligence techniques are entering real world domains, such as medicine, industrial diagnosis, communications, planning, financial forecasting and scheduling. These applications have revealed a great need for powerful methods for knowledge representation. In particular, the evolutionary nature of these domains requires a representation that takes into account temporal information. The exact timing information for things like lab-test results, occurrence of symptoms, observations, measures, as well as faults, can be crucial in these kinds of applications.

For example, in medicine, representing and reasoning about time is crucial for many tasks like prevention, diagnosis, therapeutic management, prognosis and discovery (Aliferis and Cooper 1996; Santos 1996; Hanks et al., 1995). In industrial diagnosis, it is also critical for diagnosis and prediction of events and disturbances (Arroyo et al., 1996).

To model temporal relations is a complex task. Temporal models are more complex than atemporal ones (Horvitz y Seiver 1997). Even when they involve a few variables, in a temporal model each variable and its relationships with other variables must be examined over multiple points of time. These tasks often entail an inordinate amount of computation due to the size and the complexity of the resulting model. In the context of intelligent systems, a temporal model must be capable of reasoning about the present, past and future state of the domain.

Aside from temporal considerations, real world information is usually imprecise, incomplete and noisy. The temporal model must be able to deal with uncertainty. Among the formalism proposed for dealing with uncertainty, one of the most used techniques for the development of intelligent systems are Bayesian networks (BN) (Pearl, 1988). Although BN were not designed to model temporal aspects explicitly, recently Bayesian networks have been applied to temporal reasoning under uncertainty (Kanazawa, 1991; Nicholson and Brady, 1994; Santos and Young, 1996; Aliferis and Cooper, 1996). The extension of Bayesian networks semantics to deal with temporal relationships can be complicated. The main problem is to represent each node with its dependence relationships over multiple points of time.

Although dynamic Bayesian networks are an alternative representation for these domains,

they are too complex for realistic applications. Based on the fact that in many cases there are few state changes in the temporal range of interest, in this paper we present an alternative representation called Temporal Nodes Bayesian Network (TNBN). In a TNBN each node represents an event or state change of a variable, and an arc corresponds to a causal-temporal relation. A temporal node represents the time that a variable changes state, including an option of no-change. Including more temporal nodes can represent more than one change for the same variable. The temporal intervals can differ in number and size for each temporal node, so this allows multiple granularity. Temporal information is relative, that is, there is not absolute temporal reference. We developed a mechanism for transforming the relative times to absolute, based on the timing of the observations. With this representation we can model complex real-world systems with a simple network, and use standard probability propagation techniques for diagnosis and prediction. The proposed approach is applied to the diagnosis and prediction of disturbances in power plants.

The rest of this paper is organized as follows. Section 2 introduces our approach, which is contrasted with a dynamic Bayesian network for a simple medical example. Section 3 presents a formal definition of the proposed model. Section 4 presents the inference mechanism for a TNBN. In section 5 an empirical evaluation is presented for a subsystem of a fossil power plant. Section 6 presents a brief discussion of several extensions of BNs for time modeling. Finally, section 7 presents the conclusions and future research.

## 2. TNBN: AN EXAMPLE

To illustrate the proposed temporal probabilistic model, we present the hypothetical example of the consequences of an automobile accident based on (Hanks et al., 95). The example expresses the necessity for representing temporal relations for diagnostic tasks.

Assume that at time t=0 an automobile accident occurs. The driver is a healthy 45 years old man. Contact with the steering wheel is noted. These kinds of accidents can be classified as *severe*, *moderate* or *mild*. The immediate consequences in this sort of accident are injuries to the head, abdominal cavity and internal organs, chest and extremities. For demonstration purposes we only consider *head* and *chest* injuries. Injury of the head can bruise the brain, which will cause it to begin to swell. Chest injuries can include a fractured sternum, one or both punctured lungs, and bleeding in the chest cavity. These instantaneous state changes can initiate a set of internal changes that will generate subsequent changes. For example, brain trauma will cause the brain to begin swelling. This increase of the brain volume tends to increase intracranial pressure, which in turn eventually causes *dilated pupils*, *destabilized vital signs* (pulse and blood pressure) and loss of consciousness. Bleeding into the chest cavity decreases blood volume over time, which also tends to destabilize vital signs. Internal bleeding will also eventually increase pressure on the heart, decreasing its efficiency, further destabilizing vital signs. The collision itself can be modeled as an external event, which can immediately cause certain changes in the patient's state: trauma to the brain, broken sternum, punctured lung, and bleeding in the chest cavity. These changes cause internal changes, which are not immediate: dilated pupils and unstable vital signs, depending on the severity of the accident. Suppose that we gathered the following statistics about the accidents that occurred in a specific city:

- 36.80% of the collisions (**C**) are severe, 39.20% are moderate and 24% are mild.
- If the accident is mild, the probability that head injury occurs is 0.1. The probability of an injury resulting in slight internal bleeding (**IB**) is 0.6 and gross is 0.05.
- If the accident is moderate, the probability that head injury occurs is 0.4 and the probability of an injury resulting in slight internal bleeding is 0.15 and gross is 0.65.
- If the accident is severe, the probability that head injury occurs is 0.9 and the probability of an injury resulting in slight interval bleeding is 0.4 and gross is 0.5.

This information indicates that there is a strong causal relationship between the severity of the accident and the immediate effect to the patient's state. Additionally, an expert medic provides some important temporal information about the relation between the instantaneous consequences and the symptoms (dilated pupils and unstable vitals signs).

- If a head injury (**HI**) occurs, the brain will start to swell, and if left unchecked, the swelling will cause the pupils to dilate (**PD**) within 0 to 10 minutes.
- If internal bleeding (**IB**) begins, the blood volume will start to fall, which will tend to destabilize vital signs (**VS**). The time required to destabilize signs will depend on the severity of bleeding:
  - If the bleeding is gross, it will take from 10 to 30 minutes.
  - If the bleeding is slight, it will take between 30 to 60 minutes.
- A head injury (**HI**) also tends to destabilize vital signs, taking between 0 to 10 minutes to make them unstable.

In this example there is an external event: the collision (**C**); that generates two immediate effects in the patient's state: head injury (**HI**) and internal bleeding (**IB**). These internal events produce certain posterior endogenous changes in the patient. These changes are not immediate and are manifested through two observations: *dilated pupils* (**PD**) and *unstable vital signs* (**VS**).

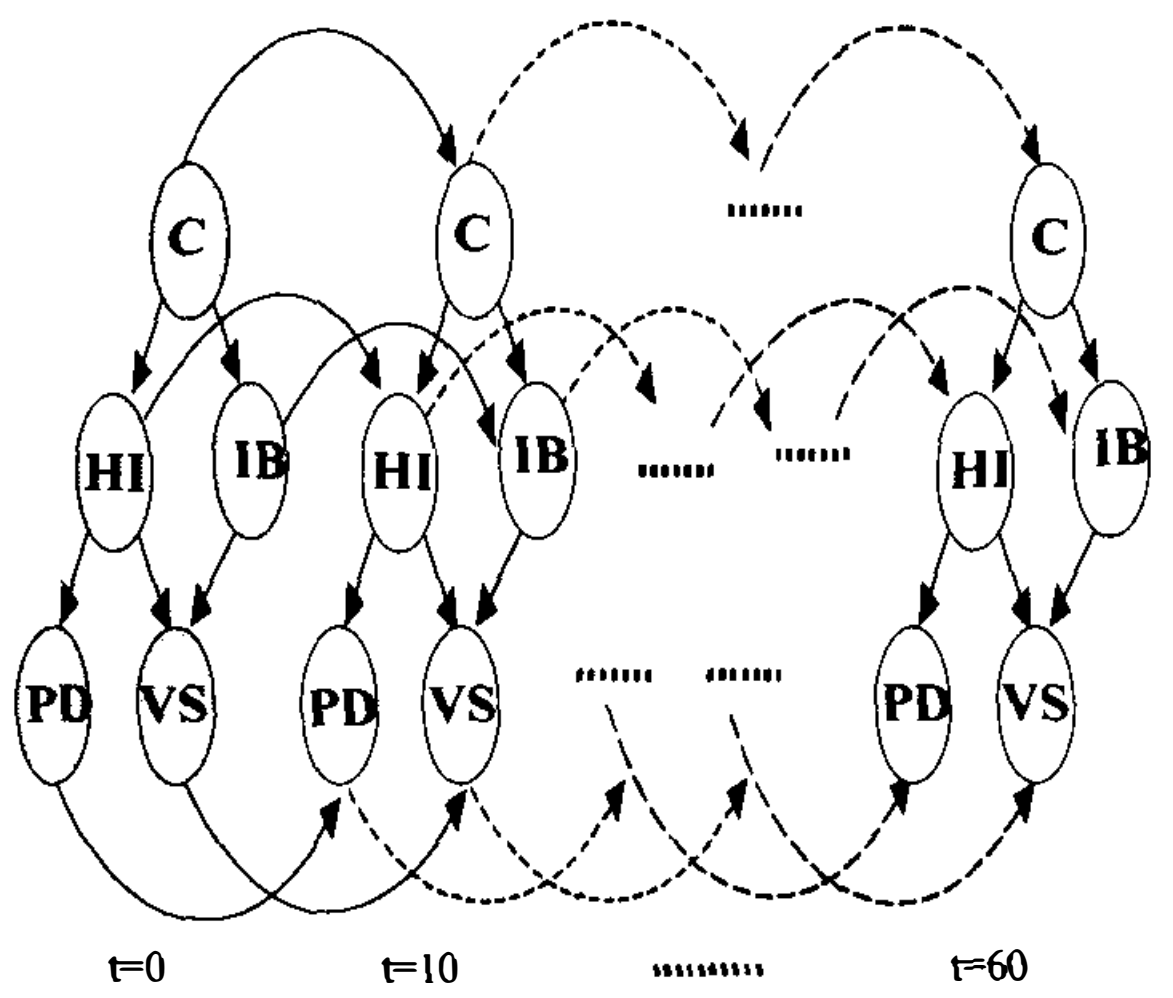

Figure 1. Dynamic Bayesian network for the "accident" example.

This information can be represented with a dynamic Bayesian network (Nicholson and Brady 1994) (DBN). A DBN is arranged into "time slices" representing the system's complete state at a single point in time. Time slices are duplicated over a predetermined time grid representing the temporal range of interest. Temporal relations are represented with arcs between nodes of different slices. Figure 1 shows a simple DBN for the "accident" example. In this case, we require a slice each 10 minutes, the maximum common divisor of the time intervals.

This is a simple DBN for the example, which considers the following assumptions: (i) a state depends only on the previous one (Markovian assumption), (ii) there are links only between the same variable at different slices. Even with these simplifications, it is a complex model in terms of storage requirements and computation time for probability propagation. If we consider a more complex model, relaxing the previous assumptions, it could become prohibitive for realistic applications. Also, the acquisition of the model (structure and parameters) could become a problem. In many cases, there are few state changes in the temporal interval of interest, called *events*. The timing of these events is usually important for diagnosis and prediction. For instance, in the medical example, the time when "unstable vital signs" and "dilated pupils" occur is crucial for the accident diagnosis. To model these changes, we require a representation of events. We propose an alternative temporal representation based on events and its time interval of occurrence.

A *Temporal Nodes Bayesian Network* (TNBN) is a Bayesian network in which each node represents an event or state change of a variable, and an arc corresponds to a causal-temporal relation. A *temporal node* represents a possible state change of a variable and the time when it happens. Each value of a temporal node is defined by an ordered pair: the value of the variable to which it changes and the time interval of its occurrence. Time intervals represent relative times between the parent events and the corresponding state change. A temporal node has an initial or default state, so a value is associated to this state with a maximum time interval (*temporal range*) and it indicates the condition of no change. Some nodes can be considered as instantaneous events, that occur or not; for these nodes temporal intervals are not defined.

In the accident example, there are 3 instantaneous events: *collision, head injury* and *internal bleeding*; and two events that can be represented by nodes with temporal intervals: *dilated pupils* and *unstable vital signs*. ***PD*** has *normal* as an initial state, and can change to *dilated* in 2 temporal intervals ([0,3],[3,5]); while ***VS*** has *normal* as an initial state, and can change to *unstable* in 3 different time intervals ([0-10],[10-30],[30-60]). Both variables have as a value the default state with an associated time interval, these correspond to the no change condition. The time intervals were defined based on the temporal information of the accident example. Figure 2 shows a TNBN model for the "accident" example, including the definition and temporal intervals for each node.

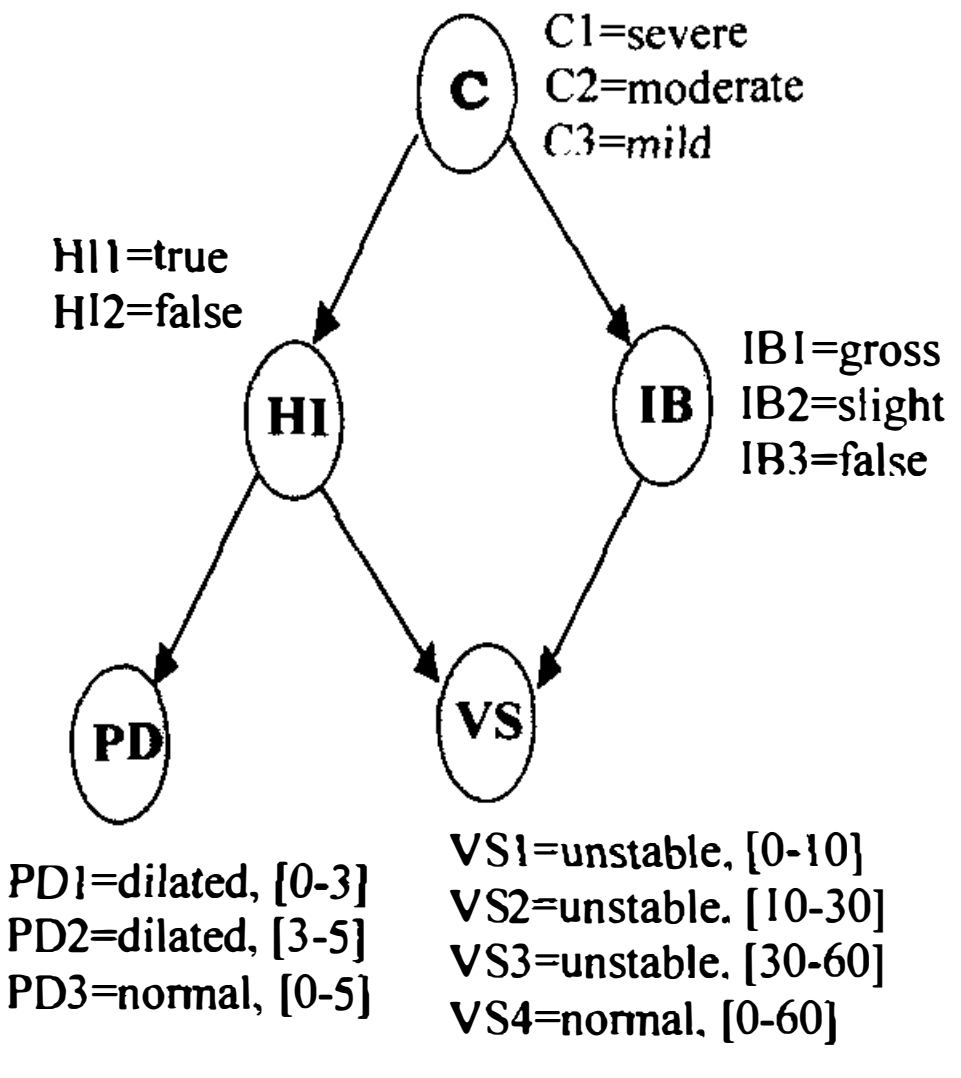

Figure 2. TNBN for the accident example.

This example illustrates that using a TNBN representation a simple and transparent model can be built for dynamic domains, in which temporal relations are important for diagnosis and prediction. In this case, the TNBN can be used, for example, to predict the consequences of an accident or to diagnose its severity. The resulting network is less complex than the corresponding DBN. The main difference with a DBN is that the representation is based on state changes at different times instead of state values.

The example informally introduces some important aspects of the TNBN model. The temporal intervals can differ in number and size for each temporal node, so this allows multiple granularity. A formal definition of a TNBN is presented in the next section.

## 3. MODEL DEFINITION

A TNBN is based on the definition of a *temporal node*. A temporal node is defined by a set of states. Each state is defined by an ordered pair: the value of the variable (to which it changes) and a time interval associated to the change in value of the variable. A temporal node is defined as follows:

> **Definition 1.** A temporal node (***TN***) is defined by a set of states, each defined by an ordered pair $(\sigma,\tau)$, where $\sigma$ is a value of a random variable and $\tau$ is the time interval associate to the change of variable value. There is a default state of no change that corresponds to the initial value (generally the "normal" value), associated to the temporal range of the node.

The values of each **TN** can be seen as the "cross product" between the set of values $(\Sigma)$ and the set of time intervals (T), except for the default state, which is associated to only one interval. The definition of the **TNs** for the accident example was presented in fig. 2.

**TNs** are connected by edges. Each edge represents a causal-temporal relationship between **TNs**. The conditional probability distribution for each node is defined as the probability of each ordered pair $(\sigma_i,\tau_i)$ given the ordered pairs of its parents $(\sigma_j,\tau_j)$. As a **TN** is defined by a set of time intervals, we can relate these time intervals based on Allen's temporal algebra (Allen 83). A temporal relationship between the time intervals of a TN is defined as:

> **Definition 2.** In a **TN** the definition of the default state is associated to temporal range of interest, **TR**. The possible temporal relationships between **TR** with the time intervals, **Ti**, of a node are: *start* (**si**), *during* (**di**) and *finish* (**fi**): **TR {si,di,fi} Ti**. The temporal relationship between each pair of time intervals is *meet* (**m**): **Ti {m} Tj.**

For example, for the temporal node *vitals signs*, the relationships between its time intervals are the following: $TR$=|0-60|, $T_1$=|0-10|, $T_2$=|10-30|, $T_3$=|30-60|; $TR$ {si} $T_1$, $TR$ {di} $T_2$, $TR$ {fi} $T_3$, $T_1$ {m} $T_2$, $T_2$ {m} $T_3$.

Finally, a Temporal Node Bayesian Network (**TNBN**) is defined as:

> **Definition 3.** A **TNBN** is defined as **TNBN**=(**V, E**), where **V** is the set of temporal nodes and **E** is the set of edge. Each temporal node is defined by an ordered pair $(\sigma, \tau)$ and the conditional probability matrix that specifies the probability of each ordered pair given its parents.

In each temporal node, the temporal intervals are relative to the parent nodes; that is, there is not an absolute temporal reference. This makes the representation more general; but, for its application, we need to associate these relative times to the actual or absolute times of the observed events. We developed a mechanism for transforming the relative times to absolute, based on the timing of the observations. In the next section we present the inference mechanism.

## 4. INFERENCE MECHANISM

As we mentioned before, the temporal intervals in each node are relative, that is, there is not absolute timing of the events until one is observed. When an initial event is detected, its time of occurrence "fixes" temporally the network. The timing of the observation is used as temporal reference for the other events. This means that the actual timing of the events represented in the network is dynamic (they are not fixed to an absolute time). For definition of the inference mechanism, we need to define some additional parameters:

**Tc** (*real time of occurrence*): is the actual time when an event is detected. As the net does not have any temporal reference, the time of occurrence of the initial event fixes temporally the network.

$\alpha$ (*real time occurrence function*): is the absolute value of the difference between the time of occurrence of a pair of events, $\alpha = |tc_i - tc_{ii}|$, where $tc_i$ is the time of occurrence of the first event and $tc_{ii}$ is the time of occurrence of the second event.

These parameters are used by the inference mechanism for determining the actual time intervals of occurrence of each event. The mechanism consists of 3 basic steps, which are as follows.

**Step 1. Event detection and time interval definition**

When an initial event is detected, its time of occurrence, "$tc_{initial}$", is utilized as temporal reference for the network. There are two possible cases, depending on the position of the *initial* node in the network: (a) the initial event corresponds to a root node, (b) the initial event corresponds to an intermediate or leaf node.

**1-(a).** In the first case, the actual value of the node can be determined (root nodes are always instantaneous events).

**1-(b).** For the second case, it is not possible to determine the value of the variable, because the event could be associated to any time interval for the state. It is necessary to wait for a second observation to determine the interval. When the next event is detected, its time of occurrence, $tc_{posterior}$, is utilized for definition of the time interval associated with the real time occurrence function, $\alpha = |tc_{initial} - tc_{posterior}|$. The value of $\alpha$ is used to set the time interval of the child node considering the parent node as the initial event. This step is applied recursively to subsequent events.

**Step 2. Propagation of the evidence**

Once the value of a node is obtained (time interval and associated state), the next step is to propagate the effect of this value through the network to update the probabilities of other temporal nodes.

**Step 3. Determination of the past and future events**

With the posterior probabilities, we can estimate the potentially past and future events based on the probability distribution of each temporal node. If there is not enough information, for instance there is only one observed event that corresponds to an intermediate node, the mechanism handles different *scenarios*. The node is instantiated to all the intervals corresponding to the observed state, and the posterior probabilities of the other nodes are obtained for each scenario. These scenarios could be used by an operator or a higher level system as a set of possible alternatives, which will be reduced when another event occurs (see [Arroyo, 1999] for details).

## 5. EXPERIMENTAL RESULTS

The proposed representation and inference mechanism is applied for fault diagnosis and prediction in a subsystem of a fossil power plant. We consider the drum level control system with four potential disturbances: a power load increase (**LI**); a feedwater pump failure (**FWPF**); a feedwater valve failure (**FWVF**); and the spray water valve failure (**SWVF**). The drum is a subsystem of a fossil power plant that provides steam to the superheater and water to the water wall of a steam generator. The drum system is composed of three systems: feedwater, water steam generator and superheater steam system. One of the main problems in the drum is to maintain the level in safe operation.

In the process, a signal exceeding its specified limit of normal functioning is called an *event* and a sequence of events that have the same underlying cause are considered as a *disturbance*. To determinate which of the disturbances is present is a complicated task, because there are similar sequences of events for the four main disturbances. We need additional information in order to determine which is the real cause. In particular, the temporal information about the occurrence of each event is important for an accurate diagnosis. For example, a feedwater flow increase (FWF) can be caused by two different events: the feedwater pump current augmentation (FWP) and feedwater valve opening increase (FWV). We can use the time difference between the occurrence of each event, FWV-FWF and FWP-FWF, for selecting the "cause" of the increase of the FW flow.

According to the process data, the time interval between a pump current augmentation and an increase of the flow (FWP-FWF) is from 25 to 114 minutes. The time interval between the valve opening increase and an increase of the flow (FWV-FWF) is from 114 to 248 seconds. Hence, if the flow increase occurs in the first time interval, the probable cause is an augmentation of the pump; but if the flow increase occurs in the second time interval, the probable cause is a valve opening increase. Figure 3 shows a TNBN that represents the events of the steam drum system of a steam generator and the definition and prior probabilities of the temporal nodes. The network structure was defined based on the knowledge of an expert operator. The definition of the time intervals for each temporal node was obtained based on knowledge about the process dynamics combined with data from a simulator.

Once the structure and time intervals were defined, the required parameters were estimated form data. The process data was generated by a full scale simulator of a fossil power plant. We selected 80% of this data-base (800 registers) for parameter learning and 20% (200 registers) for evaluation. The model was evaluated empirically using two scores: accuracy and a measure based on the Brier score (total square error). The Brier score is defined as: $BS = \Sigma^{n}_{i=1} (1 - P_i)^2$. $P_i$ is the marginal posterior probability of the correct value of each node given the evidence. The maximum Brier score is: $BS_{MAX} = \Sigma^{n} (1)^2$. A relative Brier score is defined as:

$$RBS\ (\text{in } \%) = \{1 - (BS / BS_{MAX})\} \times 100$$

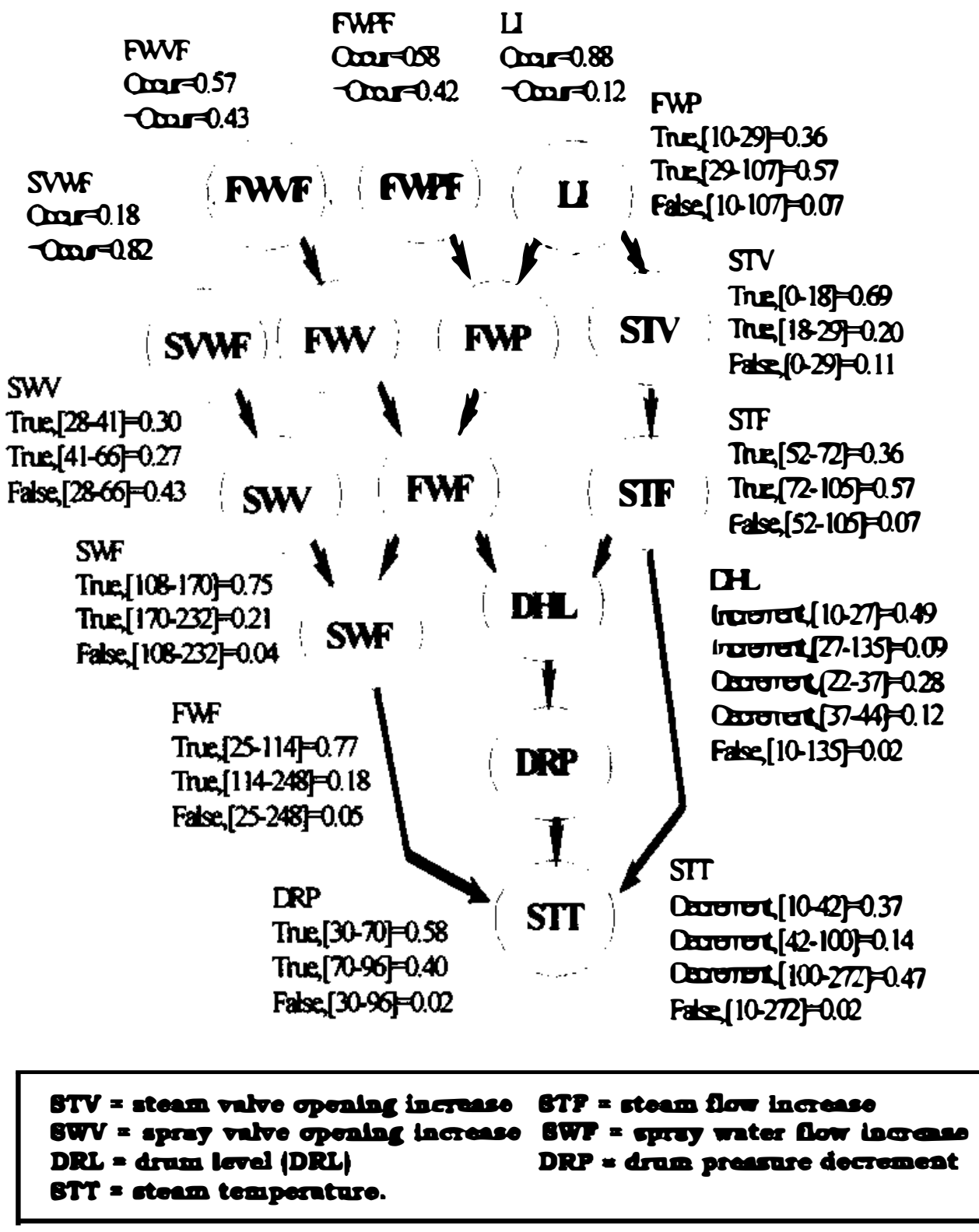

Figure 3. TNBN for the steam drum system.

The test methodology includes three basic steps: (i) assign a value to a subset of nodes, (ii) propagate the evidence, and (iii) compare the posterior probabilities of the nodes with the actual values. The assigned nodes were selected for 3 sets of tests: (1) *Prediction*: root nodes are observed (LI, FWPF, FWVF and SWVF); (2) *Diagnosis*: leaf nodes are observed (STT, STF and SWF); and (3) *Prediction and diagnosis*: intermediate nodes are observed (STV, FWP, FWV and SWV).

Table 1 shows the results of the evaluation for the three sets of tests in terms of the mean and the standard deviation for both scores. These results show the prediction and diagnosis capacity of the temporal model in a real process. Both scores are between 80 and 97% for all the set of tests, with better results when intermediate nodes are observed, and slightly better results for prediction compared to diagnosis. We consider that these differences have to do with the "distance" between assigned and unknown nodes, and with the way that the temporal intervals were defined. We are encouraged by the fact that the model produce a reasonable accuracy in times that are compatible with real-time decision making.

Table 1. Empirical evaluations results.

| Parameter | μ | σ |
|---|---|---|
| Prediction<br>% of RBS<br>% of Accuracy | <br>87.37<br>84.48 | <br>9.19<br>14.98 |
| Diagnosis<br>% of RBS<br>% of Accuracy | <br>84.25<br>80.00 | <br>8.09<br>11.85 |
| Diagnosis and Prediction<br>% of RBS<br>% of Accuracy | <br>95.85<br>94.92 | <br>4.71<br>8.59 |

## 6. RELATED WORK

In this section, we review related work in temporal Bayesian networks and contrast it with our approach. The "time net" of Kanazawa [Kanazawa, 1991] is a kind of Bayesian network with a formal declarative language of random variables for making inferences. The events are considered to occur at an instant of time while facts are considered to occur over a series of time points. The "Dynamic Belief Networks" of Nicholson and Brady [1994] consider a DBN where the network has certain structure at time *ti* and a different structure at time *ti+1*. The DBN is built dynamically, reflecting the dynamic changes in the environment. The "Network of exogenous events and endogenous changes" [Hanks et al, 1995] is a probabilistic model for reasoning about the system as it changes over time, both due to exogenous events and endogenous changes. An exogenous event generally refers to an instantaneous change in the process state. Endogenous changes are modeled using a local inference model, a simple arbitrary linear model.

Santos and Young [1996] proposed a probabilistic temporal network (PTN). Bayesian networks provide the probabilistic basis for the management of uncertainty. Allen's interval algebra and its thirteen relations provide the temporal basis [Allen, 83]. The nodes of the network are temporal aggregates and the edges are the causal/temporal relationships between aggregates. Each aggregate represents a process changing over time. The temporal aggregates are temporal random variables, defined by an ordered pair (random variable plus Allen's intervals).

Aliferis and Cooper [1996] proposed an extension of Bayesian networks called "Modificable Temporal Bayesian Networks with Single-granularity (MTBN-SG). A MTBN-SG is an extended time-sliced Bayesian network defined over a range of time points. The temporal graph is a directed graph (possibly cyclic) composed of nodes and arcs corresponding to 3 types of variables; ordinary, mechanism and time-lag quantifier variable. The resulting graph can have cycles to allow expressions of recurrence and feedback. This model has great capacity for representing temporal and atemporal information. The problems with this model are that the joint probability distribution is not compatible with the Bayesian model and that it only supports a single granularity for the size of the time step. Extending the model to support multiple granularity appears problematic, it is difficult to combine two events with different time ranges. Also, the acquisition of the quantitative information appears to be a big problem because of the excessive number of probabilities required.

In summary, previous probabilistic temporal models are, in general, based on static models repeated at different times. The resulting models are quite complex for many applications, so they do not satisfy the knowledge acquisition and computational tractability criteria [Aliferis and Cooper, 1996].

These models support a single granularity. In contrast, the TNBN model is based on representing changes of state in each node. If the number of possible state changes for each variable in the temporal range is small, the resulting model is much simpler. This facilitates temporal knowledge acquisition and allows efficient inference using probability propagation. The model supports granularity, with different numbers of temporal intervals for each node, and different durations for each interval within a node.

## 7. CONCLUSIONS

This paper presents a novel approach for dealing with uncertainty and time called Temporal Nodes Bayesian Network. A TNBN is an extension of Bayesian networks for dealing with temporal information. Each event or state change of a variable is associated with a time interval. The definition of the number of time intervals and their duration for each node is free (multiple granularity). Temporal reasoning is based on probability propagation and gives the time of occurrence of events or state changes with some probability value. The main difference with previous probabilistic temporal models is that the representation is based on state changes instead of state values at different times. This makes the model much simpler in many applications in which there are few changes for each variable in the temporal range of interest. Temporal information in a TNBN is relative, that is, there is not absolute temporal reference. We developed a mechanism for transforming the relative times to absolute, based on the timing of the observations. An empirical evaluation is presented for a complex real problem, a drum level system of a fossil power plant, in which this approach is used for fault diagnosis and event prediction with good results.

Our future work will focus on validating our approach with additional experiments on other real-world domains. Also, we will incorporate qualitative temporal constraints that could facilitate knowledge acquisition and checking the temporal consistency of the model.

**Acknowledgments**

Thanks to Eduardo Morales, Brent Petersen and the anonymous referees for their comments which improved this article.